\documentclass[10pt,twocolumn,letterpaper]{article}

\usepackage{iccv}
\usepackage{times}
\usepackage{epsfig}
\usepackage{graphicx}
\usepackage{amsmath}
\usepackage{amssymb}
\usepackage{multirow}
\usepackage[ruled,vlined]{algorithm2e}
\usepackage{algorithmic}
\usepackage{bbm}

\usepackage{pdfpages}
\usepackage[pagebackref=true,breaklinks=true,letterpaper=true,colorlinks,bookmarks=false]{hyperref}

\DeclareMathOperator*{\argmax}{argmax}

\DeclareMathOperator{\BigO}{O}

\iccvfinalcopy % *** Uncomment this line for the final submission

\def\iccvPaperID{276} % *** Enter the ICCV Paper ID here

\ificcvfinal\fi
\begin{document}

%%%%%%%%% TITLE
\title{Merge or Not? Learning to Group Faces via Imitation Learning}

\author{Yue He\\
SenseTime\\
{\tt\small heyue@sensetime.com}
\and
Kaidi Cao\\
SenseTime\\
{\tt\small caokaidi@sensetime.com}
\and
Cheng Li\\
SenseTime\\
{\tt\small chengli@sensetime.com}
\and
Chen Chang Loy\\
The Chinese University of Hong Kong\\
{\tt\small ccloy@ie.cuhk.edu.hk }
}

\maketitle
%\thispagestyle{empty}

%%%%%%%%% ABSTRACT
\begin{abstract}

Given a large number of unlabeled face images, face grouping aims at clustering the images into individual identities present in the data. This task remains a challenging problem despite the remarkable capability of deep learning approaches in learning face representation. In particular, grouping results can still be egregious given profile faces and a large number of uninteresting faces and noisy detections. Often, a user needs to correct the erroneous grouping manually. In this study, we formulate a novel face grouping framework that learns clustering strategy from ground-truth simulated behavior. This is achieved through imitation learning (a.k.a apprenticeship learning or learning by watching) via inverse reinforcement learning (IRL). In contrast to existing clustering approaches that group instances by similarity, our framework makes sequential decision to dynamically decide when to merge two face instances/groups driven by short- and long-term rewards.
Extensive experiments on three benchmark datasets show that our framework outperforms unsupervised and supervised baselines.
\end{abstract}
\vspace{-0.6cm}

%%%%%%%%%%%%%%%%%%%%%%%%%%%%%%%%%%%%%%%%%%%%%%%%%%%%%%%%%%%%%%%%%%%%%%%%%%%%%%%%%
\section{Introduction}
\label{sec:introduction}
%%%%%%%%%%%%%%%%%%%%%%%%%%%%%%%%%%%%%%%%%%%%%%%%%%%%%%%%%%%%%%%%%%%%%%%%%%%%%%%%%

Face grouping is an actively researched computer vision problem due to its enormous potential in commercial applications. It not only allows users to organize and tag photos based on faces but also retrieve and revisit huge quantity of relevant images effortlessly.

\begin{figure}[t]
\begin{center}
   \includegraphics[width=0.95\linewidth]{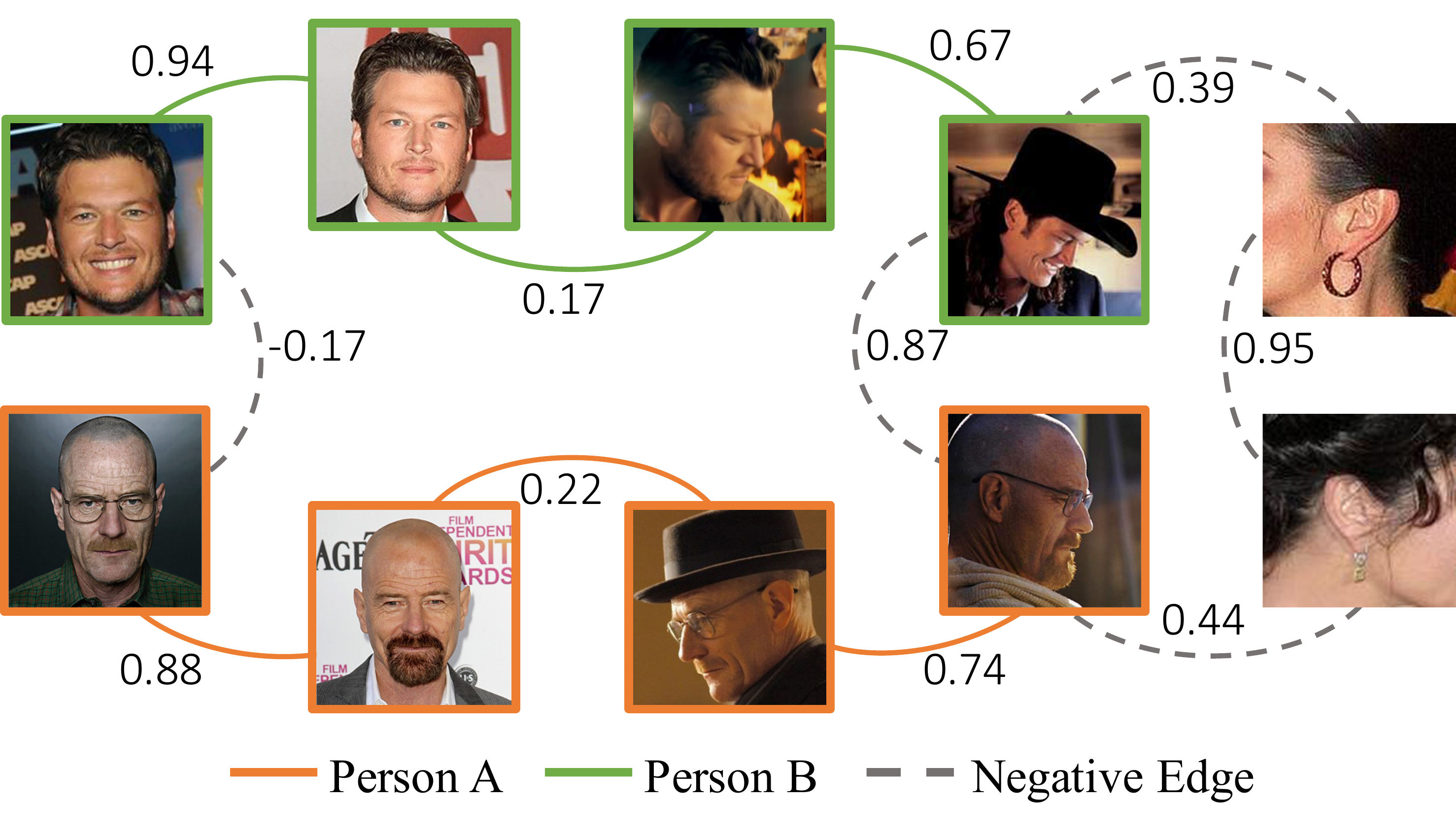}
\end{center}
\vskip -0.45cm
   \caption{\textbf{Cosine angle in a deep feature space.} We measure the cosine angle between the deep feature vector of two faces. It is noted that even for two men with significantly different appearances, the angle between their profiles and noise faces (gray dash lines with 0.39$\sim$0.44) is much larger than one's frontal and his own profile (0.17 and 0.22).}
\label{fig:profileSimilairty}
\end{figure}

The performance of face grouping significantly benefits from the recent emergence of deep learning approaches~\cite{chen2016unconstrained,parkhi2015deep,schroff2015facenet,sun2014deep,taigman2014deepface,wen2016discriminative}. Nevertheless, we still observe some challenges when we apply existing methods on real-world photo albums. In particular, we found that deeply learned representation can still perform poorly given \textit{profile faces and false detections}. In addition, there is no obvious mechanism to disambiguate \textit{large quantity of non-interested faces}\footnote{Non-interested faces refer to faces that we do not want to group (\eg~faces in the background). This is the term popularized by the earlier work in face clustering~\cite{zhu2011rank}.} that are captured under the same condition with the person of interests. 
We provide an illustrative example in Fig.~\ref{fig:profileSimilairty}, of which results were obtained from the Inception-v3 model~\cite{szegedy2015rethinking} fine-tuned with MS-Celeb-1M~\cite{guo2016msceleb} images with face identity.
Despite the model achieves an accuracy of 99.27\% on LFW~\cite{huang2007labeled}, which is on par with the accuracy reported by a state-of-the-art method~\cite{wen2016discriminative}, its performance on the open-world face grouping task is unsatisfactory. 
We attempted to adapt the deep model with open-world albums~\cite{zhang2016joint} but with limited success. We show experimental results in Sec.~\ref{sec:experiments}. Learning such an open-world model is still far from being solved due to highly imbalanced data (much more frontal faces compared to profile instances in existing datasets) and a large negative space to cover.

Thinking about humans, we tend to execute a visual grouping task in sequence with intermediate decision to govern our next step, like playing a jigsaw puzzle~\cite{xie2008tangibles} with pieces of varying visual complexity. First we will link pieces with strong correlation and high confidence, then gain insights and accumulate visual evidence from these stable clusters. Consequently, a larger group can be formed through merging ambiguous positives and discarding uninteresting outliers. In the process, we may exploit contextual cues and global picture considering other samples.

The above intuition motivates a novel face grouping framework. Our goal is not to design a better deep representation, but learning to make better merging/not-merging decision from expert?s demonstration using existing representation.
In particular, we wish to introduce intermediate sequential decision between the clustering steps, \ie, when to merge two samples or groups given the dynamic context.
Towards this goal, we assume different clustering states, where the states differ in their current partitions of data. At each time step, an agent will choose from two possible actions, \ie, to merge or not to merge a pair of face groups. The process responds at the next time step by moving to a new state and provides a reward to the agent. A sequence of good actions would lead to higher accumulative reward than suboptimal decisions.

Learning a decision strategy in our problem is non-trivial. In particular, the decision process is adversely affected by uninteresting faces and noisy detections. Defining a reward function for face grouping is thus not straightforward, which needs to consider the similarity of faces, group consistency, and quality of images.
In addition, we also need to consider the operation cost involved, \ie, the manual human effort spent on adding or removing a photo from a group.
It is hard to determine the relative weights of these terms a-priori. This is in contrast to (first person) imitation learning setting of which the reward is usually assumed known and fixed, \eg, using the change of game score~\cite{mnih2015human}.

\noindent
\textbf{Contributions:}
We make the following contributions to overcome the aforementioned challenges:

\noindent
1) We formulate a novel face grouping framework based on imitation learning (IL)
via inverse reinforcement learning~\cite{ng2000algorithms,ross2011reduction}. 
To our knowledge, this is the first attempt to address visual clustering via inverse reinforcement learning. Once learned, the policy can be transferred to unseen photo albums with good generalization performance.

\noindent
2) We assume the reward as an unknown to be ascertained through learning by watching an expert's behavior. We formulate the learning such that both short- and long-terms rewards are considered. The formal considers similarity, consistency and quality of local candidate clusters; whereas the latter measures the operation cost to get from an arbitrary photos partition to the final ground-truth partition. The new reward system effectively handles the challenges of profile, noisy, and uninteresting faces, and works well with conventional face similarity under an open-world context.

\noindent
3) We introduce a large-scale dataset called Grouping Faces in the Wild (GFW) to facilitate the research of real-world photo grouping. The new dataset contains $78,000$ faces of $3,132$ identities collected from a social network. This dataset is realistic, providing a large number of uninteresting faces and noisy detections.

Extensive experiments are conducted on three datasets, namely, LFW simulated albums, ACCIO dataset (Harry Potter movie)~\cite{esam2015accio}, and the GFW introduced by us.
We show that the proposed method can be adapted to a variety of clustering algorithms, from the conventional k-means and hierarchical clustering to the more elaborated graph degree linkage (GDL) approach~\cite{zhang2012graph}.
We show that it outperforms a number of unsupervised and supervised baselines.

%\noindent \textbf{Open-source:} Our codes and data will be released to facilitate future studies.

%%%%%%%%%%%%%%%%%%%%%%%%%%%%%%%%%%%%%%%%%%%%%%%%%%%%%%%%%%%%%%%%%%%%%%%%%%%%%%%%%
\section{Related Work}
\label{sec:related_work}
%%%%%%%%%%%%%%%%%%%%%%%%%%%%%%%%%%%%%%%%%%%%%%%%%%%%%%%%%%%%%%%%%%%%%%%%%%%%%%%%%
\vspace{-0.1cm}
\noindent
\textbf{Face Grouping:}
Traditional face clustering methods~\cite{cao2015diversity,li2004bayesian,otto2015efficient,zhu2011rank} are usually purely data-driven and unsupervised. They mainly focus on finding good distance metric between faces or effective subspaces for face representation. For instance, Zhu \etal~\cite{zhu2011rank} propose a rank-order distance that measures the similarity between two faces using their neighboring information. Fitzgibbon and Zisserman~\cite{fitzgibbon2003joint} further develop a joint manifold distance (JMD) that measures the distance between two subspaces, each of which invariant to a desired group of transformations.
Zhang~\etal~\cite{zhang2012graph} propose agglomerative clustering on a directed graph to better capture global manifold structures of face data.
There exist techniques that employ user interactions~\cite{tian2007face}, extra information on the web~\cite{berg2004names} and prior knowledge of family photo albums~\cite{xia2014face}. 
Deep representation is recently found effective for face clustering~\cite{schroff2015facenet}, and large-scale face clustering has been attempted~\cite{otto2016clustering}.
Beyond image-based clustering, most existing video-based approaches employ pairwise constraints derived from face tracklets~\cite{cinbis2011unsupervised,wu2013constrained,xiao2014weighted,zhang2016joint} or other auxiliary information~\cite{el2010face,tang2015face,zhou2015multi} to facilitate face clustering in video. The state-of-the-art method by Zhang~\etal~\cite{zhang2016joint} adapts DeepID2+ model~\cite{sun2015deeply} to a target domain with joint face representation adaptation and clustering.

In this study, we focus on image-based face grouping without temporal information. Our method differs significantly to existing methods~\cite{zhang2016joint} that cluster instances by deep representation alone. Instead, our method learns from experts to make sequential decision on grouping considering both short- and long-term rewards. It is thus capable of coping with uninteresting faces and noisy detections effectively.

\vspace{0.1cm} \noindent
\textbf{Clustering with Reinforcement Learning:}
There exist some pioneering studies that explored clustering with RL.
Likas~\cite{likas1999reinforcement} models the decision process of assigning a sample from a data stream to a prototype, \eg, cluster centers produced by on-line K-means.
Barbakh and Fyfe~\cite{barbakh2007clustering} employ RL to select a better initialization for K-means. 
Our work differs to the aforementioned studies: (1) \cite{barbakh2007clustering,likas1999reinforcement} are unsupervised, \eg, their loss is related to the distance from data to a cluster prototype. In contrast, our framework guides an agent with a teacher's behavior. (2) We consider a decision that extends more flexibly to merge arbitrary instances or groups. We also investigate a novel reward function and new mechanisms to deal with noises.

\vspace{0.1cm} \noindent
\textbf{Imitation Learning:} Ng and Russel~\cite{ng2000algorithms} introduced the concept of \textit{inverse reinforcement learning} (IRL), which is also known as \textit{imitation learning} or apprenticeship learning~\cite{abbeel2004apprenticeship}. The goal of IRL is to find a reward function to explain observed behavior of an expert who acts according to an unknown policy. Inverse reinforcement learning is useful when a reward function is multivariate, \ie, consists of several reward terms of which the relative weights of these terms are unknown a-priori.
Imitation learning was shown effective when the supervision of a dynamic process is obtainable, \eg, in robotic navigation \cite{abbeel2004apprenticeship}, activity understanding and forecasting~\cite{kitani2012activity} and visual tracking~\cite{xiang2015learning}.

% The applications of MDP can be found on different computer vision tasks, \eg, feature selection~\cite{lucas2005Qlearning,Karayev2014anytime}, human activity forecasting~\cite{kitani2012activity}, and interactive data annotation~\cite{russakovsky2015best}. tracking~

%%%%%%%%%%%%%%%%%%%%%%%%%%%%%%%%%%%%%%%%%%%%%%%%%%%%%%%%%%%%%%%%%%%%%%%%%%%%%%%%%
\section{Overview}
\label{sec:overview}

\begin{figure}[t]
\begin{center}
   \includegraphics[width=\linewidth]{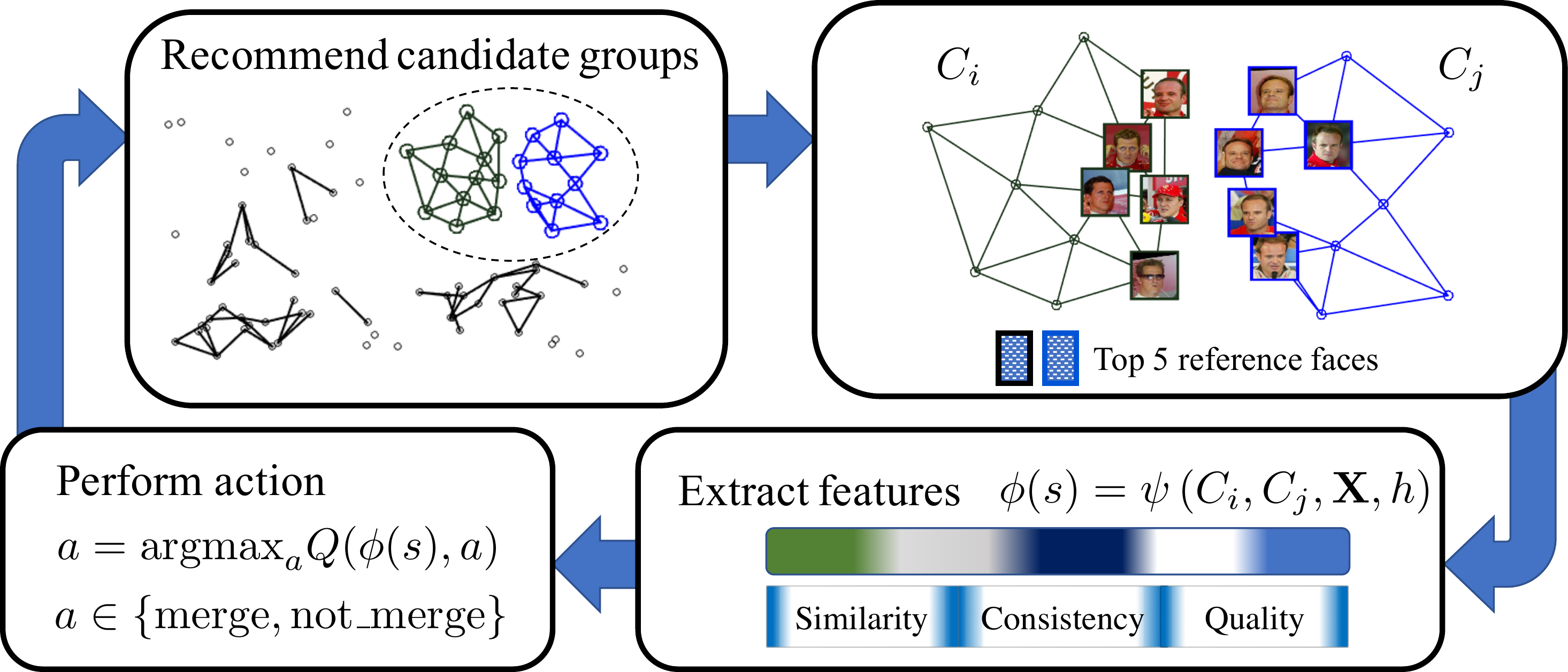}
\end{center}
\vskip -0.45cm
   \caption{Face grouping by the proposed framework.}
\label{fig:decisionIllustration}
\end{figure}

An illustration of the proposed framework is given in Fig.~\ref{fig:decisionIllustration}. We treat grouping as a sequential process. In each step during test time, two candidate groups $C_i$ and $C_j$ are chosen. Without loss of generality, a group can be formed by just a single instance. 
Given the two groups, we extract meaningful features to characterize their similarity, group consistency, and image quality. 
Based on the features, an agent will then perform an action, which can be either i) merging the two groups, or ii) not merging the two groups. Once the action is executed accordingly, the grouping proceeds to select the next pair of groups.
The merging stops when there are no further candidate groups can be chosen, \eg, the similarity between any groups is higher than a pre-defined threshold.
Next, we define some key terminologies.

\vspace{0.1cm}
\noindent 
\textbf{Recommender}:  At each time step we pick and consider the merging of two face groups. The action space is large with a complexity of $O(N^2)$, where $N$ is the number of groups. This adds hurdles to both learning and test stages.
To makes our approach scalable, we employ a recommender, $M$, which recommends two candidates cluster $C_i$ and $C_j$ at each time step. This reduces the $O(N^2)$ action space to a binary problem, \ie, to merge or not to merge a pair of face groups.
A recommender $M$ can be derived from many classic clustering algorithms especially agglomerative-based algorithm like hierarchical clustering (HC), ranked-ordered clustering~\cite{zhu2011rank} and GDL approach~\cite{zhang2012graph}.
For instance, hierarchical clustering-based $M$ always suggest two clusters that are nearest by some distance metric.
In Sec.~\ref{sec:experiments}, we perform rigorous evaluations on plausible choices of a recommender.

\vspace{0.1cm}
\noindent 
\textbf{State:} Each state $s_t = (h_t, H_t )\in \mathcal{S}$, contains the current grouping partition $h_t$ and recommender history $H_t$, at time step $t$. In each discrete state, the recommender $M$ will recommend a pair of cluster $(C_i, C_j) = M( s )$ based on the current state.

\vspace{0.1cm}
\noindent 
\textbf{Action:} An action is denoted as $a$. An agent can execute two possible actions, \ie, merge two groups or not. That is the action set is defined as $\mathcal{A} = \left\{\mathrm{merge}, \mathrm{not}\text{\_}\mathrm{merge}\right\}$, and $a\in \mathcal{A}$. %, where 1 is for accept a merging and -1 for not.

\vspace{0.1cm}
\noindent 
\textbf{Transition:}
If a merging action is executed, candidate groups $C_i$ and $C_j$ will be merged. The corresponding partition is updated as $h_{t+1} \leftarrow \{ h_t \backslash \{C_i, C_j\} \}  \cup \{ C_i \cup C_j \} $. Otherwise, the partition remains unchanged, $h_{t+1} \leftarrow h_t $. The candidate information will be appended to the history $H_{t+1}$ so that the same pair would not be recommended by $M$.
The transition is thus represented as $s_{t+1} = T(s_t,a)$, where $T(\cdot)$ denotes the transition function, and $s_{t+1} = (h_{t+1}, H_{t+1})$ and $s_t = (h_t, H_t)$.

%A local term $R_{\mathrm{merge}}$ gauge how strong two cluster should be merged. It tend to merge clusters with strong similarity and high quality. In our framework we count multiple local cues for two cluster, and use IRL to learn this reward term. 2. The global term we called $R_{\mathrm{OP}}$, we use an operator cost to measures the distance from current partition to the ground-truth partition.

%\begin{equation}
%    R = R_{\mathrm{merge}} + R_{\mathrm{operator}}
%\end{equation}

%------------------------------------------------------------------------------%
\section{Learning Face Grouping by Imitation}
\label{sec:learning_face_grouping}
%------------------------------------------------------------------------------%

The previous section explains the face grouping process at test time. An agent is used to determine the right action at each step, \ie, merging or not merging a pair of groups.
To learn an agent with the desired behavior, we assume access to demonstrations by some expert.
In our study, we obtain these demonstrations from a set training photo albums of which the ground-truth partition of the photos is known. Consequently, given any two candidate groups, $C_i$ and $C_j$, we know if merging them is a correct action or not. 
These ground-truth actions $\left\{a_\mathrm{GT}\right\}$ represent the pseudo expert's behavior.

%This ground-truth can be used to form trajectories of grouping represented as state sequences, \eg, $\left\{s_0, s_1, \dots\right\}$, which captures the expert's behavior.

%Under this setting, it is typically assumed that the purpose of demonstrations is to learn a policy, which represents the mapping from states to probability distributions over actions. 
%%
%Nevertheless, in the seminal paper by Ng and Russel on inverse reinforcement learning~\cite{ng2000algorithms}, they suggest that the reward function often provides a much more parsimonious description of behavior. Thus, they propose instead to recover the expert's reward function rather than the policy, and use the recovered reward function to generate desirable behavior. 

Towards the goal of learning an agent from the expert's behavior, we perform the learning in two stages: (1) we find a reward function to explain the behavior via inverse reinforcement learning~\cite{ng2000algorithms}, (2) with the learned reward function we find a policy that maximizes the cumulative rewards.

Formally,  let $R: \mathcal{S} \times \mathcal{A} \rightarrow \mathbb{R}$ denotes the reward function, which rewards the agent after it executes action $a$ in state $s$. And $\mathcal{T}$ is a set of state transition probabilities upon taking action $a$ in state $s$.
For any policy $\pi$, a value function $V^\pi$ evaluates the value of a state as the total amount of reward an agent can expect to accumulate over the future, starting from that state, $s_1$,
\begin{equation}
\label{eqn:valuefunction}
    V^\pi(s_1 ) = \mathrm{E} \left[ \sum\nolimits_{t = 0}^\infty \gamma^{t-1} R(s_t, a_t | \pi ) \right],
\end{equation}
where $\gamma$ is a discount factor.

An action-value function $Q^\pi$ is used to judge the value of actions, according to 
%
%\begin{equation}
%\label{eqn:Qfunction}
%Q^\pi( s, a  | \pi ) = R(s,a) + \gamma \mathrm{E}_{s' \sim \mathcal{T}(s,a)} V(s' |\pi),
%\end{equation}
\begin{equation}
\label{eqn:Qfunction}
Q^\pi( s, a ) = R(s,a) + \gamma V^\pi (s' | s' = T(s, a ) ),
\end{equation}
where the notation $s' = T(s,a)$ represents the transition to state $s'$ after taking an action $a$ at state $s$.
Our goal is to first uncover the reward function $R$ from expert's behavior, and find a policy $\pi$ that maximizes $Q^\pi(s,a)$. 
%
%At the end of the learning, we will have an action-value function $Q$ that approximates $Q^\pi$, and it can be used to ....

\vspace{0.1cm}
\noindent 
\textbf{Rewards:} In our study, the reward function that we wish to learn consists of two terms, denoted as 
\begin{equation}
\label{eqn:reward}
R = R_{\mathrm{short}} + \beta R_{\mathrm{long}}.
\end{equation}
The first and second term corresponds to short- and long-term rewards, respectively. The parameter $\beta$ helps balance the scale of the two terms.  
The short-term reward is multivariate. It considers how strong two instances/groups should be merged locally based on face similarity, group consistency, and face quality. 
A long-term reward captures more far-sighted clustering strategy through measuring the operation cost to get from an arbitrary photos partition to the final ground-truth partition. 
Note that during the test time, the long-term reward function is absorbed in our learned action-value function for a policy $\pi$, thus no ground-truth is needed during testing.
We provide explanations on the short- and long-term rewards as follows.

%------------------------------------------------------------------------------%
\subsection{Short-Term Reward}
\label{subsec:short_term}
%------------------------------------------------------------------------------%

%\subsection{Learn Merging Reward via IRL}

Before a human user decides a merge between any two face groups, he/she will determine how close the two groups are in terms of face similarity. In addition, he/she may consider the quality and consistency of images in each group to prevent any accidental merging of uninteresting faces and noisy detections. 
We wish to capture such a behavior through learning a reward function. 

The reward is considered short-term since it only examines the current groups' partition.
Specifically, we compute the similarity between two groups, the quality for each group and photos consistency in each group as a feature vector $\phi(s)$, and we project this feature into a scalar reward, 
\begin{equation}
\label{eqn:rewardShort}
    R_{\mathrm{short}}(s, a ) = y(a) \left( \mathbf{w}^\mathsf{T} \phi(s) + b \right),
\end{equation}
where $y(a)=1$ if action $a = \mathrm{merge}$, and $y(a)=-1$ if $a = \mathrm{not}\text{\_}\mathrm{merge}$.
Note that we assume the actual reward function is unknown and $(\mathbf{w},b)$ should be learned through IRL.
We observe that through IRL, a powerful reward function can be learned. An agent can achieve a competitive result even by myopically deciding based on one step's reward function rather than multiple steps. We will show that optimizing $(\mathbf{w},b)$ is equivalent to learning a hyperplane in support vector machine (SVM) (Sec.~\ref{subsec:learning}). 

Next, we describe how we design the feature vector $\phi(s)$, which determines the characteristics an agent should examine before making a group merging decision.
A feature vector is extracted considering the candidate groups, all faces' representation $\mathbf{X}$ in the groups, and current partition $h$, that is $\phi(s) = \psi \left( C_i, C_j, \mathbf{X}, h \right)$. 

The proposed feature vector contains three kinds of features, so as to capture face similarity, group consistency, and image quality. All face representation are extracted from Inception-v3 model~\cite{szegedy2015rethinking} fine-tuned with MS-Celeb-1M~\cite{guo2016msceleb}. More elaborated features can be considered given the flexibility of the framework.

\noindent
\textbf{Face Similarity:} We compute a multi-dimensional similarity vector to describe the relationship between two face groups $C_i$ and $C_j$. Specifically, we first define the distance between the representation of two arbitrary faces $\mathbf{x}^u_i \in C_i$, and $\mathbf{x}^v_j \in C_j$ as $d(\mathbf{x}^u_i,\mathbf{x}^v_j)$. The subscript on $\mathbf{x}$ indicates its group. In this study, we define the distance function as angular distance.
We then start from $C_i$: for a face $\mathbf{x}^u_i$ in $C_i$, we compute its distance to all the faces in $C_j$ and select a median from the resulting distances. %\footnote{We attempted average to replace the median operator but observed poorer performance.}. 
That is 
\begin{equation}
    d_\mathrm{med}(\mathbf{x}^u_i, C_j) = \mathrm{median} \left\{ d(\mathbf{x}^u_i , \mathbf{x}^1_j), \dots, d(\mathbf{x}^u_i , \mathbf{x}^{n_j}_j) \right\},
\end{equation}
where $n_j = |C_j|$. We select $\eta$ number of instances with the shortest distances from $\left\{ d_\mathrm{med}(\mathbf{x}^u_i, C_j), \forall u \right\}$ to define the distance from $C_i$ to $C_j$. Note that the distance is not symmetric. Hence, we repeat the above process to obtain another $\eta$ shortest distances from $\left\{ d_\mathrm{med}(C_i,\mathbf{x}^v_j), \forall v \right\}$ to define the distance from $C_i$ to $C_j$.
Lastly, these $2\eta$ distances are concatenated to form a $2\eta$-dimensional feature vector.

\noindent
\textbf{Group Consistency:} Group consistency measures how close the samples in a group to each other. Even two groups have high similarity in between their respective members, we may not want to merge them if one of the group is not consistent, which may happen when there are a number of non-interesting faces inside the group.
We define the consistency of a group as the median of pairwise distances between faces in the group itself. Given a group $C_i$:
\begin{equation}
    consistency(C_i)\!=\!\mathrm{median} \left\{ d(\mathbf{x}^u_i , \mathbf{x}^v_i), u\! \neq\!v, \forall (u,v)\!\in \!C_i \right\}.
\end{equation}
Consistency is computed for the two candidate groups, contribute a two-dimensional feature vector to $\phi(s)$.

\noindent
\textbf{Face Quality:} As depicted in Fig.~\ref{fig:profileSimilairty}, profile faces and noises could easily confuse a state-of-the-art face recognition model. To make our reward function more informed on the quality of the images, we train a linear classifier by using annotated profile and falsely detected faces as negative samples, and clear frontal faces as positive samples. A total of 100k face images extracted from movies is used for training. The output of the classifier serves as the quality measure. Here, we concatenate the quality values of the top $\eta$ faces in each of the two groups to form another $2\eta$-dimensional features to $\phi(s)$.

%------------------------------------------------------------------------------%
\subsection{Long-Term Reward}
\label{subsec:long_term}
%------------------------------------------------------------------------------%

While the short-term reward $R_\mathrm{short}$ captures how likely two groups should be merged given the current partition, the long-term reward $R_\mathrm{long}$ needs to encapsulate a more far-sighted clustering strategy. %Ideally, the long-term reward should capture how far the current partition is away from the ground-truth partition. 

To facilitate the learning of this reward, we introduce the term `\textit{operation cost}', which measures the efforts needed to manipulate the images in the current partition to approach to ground-truth partition. 
Formally, given a partition $h \in \mathcal{V}$ and ground-truth partition $g \in \mathcal{V}$. A sequence of operations $o_i \in \mathcal{O}:\mathcal{V} \rightarrow \mathcal{V}$ can be executed to gradually modify the partition $h$ to $g$. 
The cost function $c : \mathcal{O} \rightarrow \mathbb{R}_{>0}$ maps each type of operations into a positive time cost. then we define $Op(h,g)$ as the minimal cost for this change:
\begin{equation}    
\begin{aligned}
\label{eqn:operation_cost}
Op(h,g) = \min_{\Gamma, o_1 \ldots o_\Gamma} & \sum\nolimits_{t=1}^\Gamma c\left( o_t \right), \\
\mathrm{s.t.~} g & = o_\Gamma \cdot \ldots \cdot o_2 \cdot o_1 \cdot h \\
 o_t & \in \mathcal{O}
\end{aligned}
\end{equation}
where $\Gamma$ is the number of steps needed to get from $h$ to $g$.

\begin{figure}[t]
\begin{center}
   \includegraphics[width=0.95\linewidth]{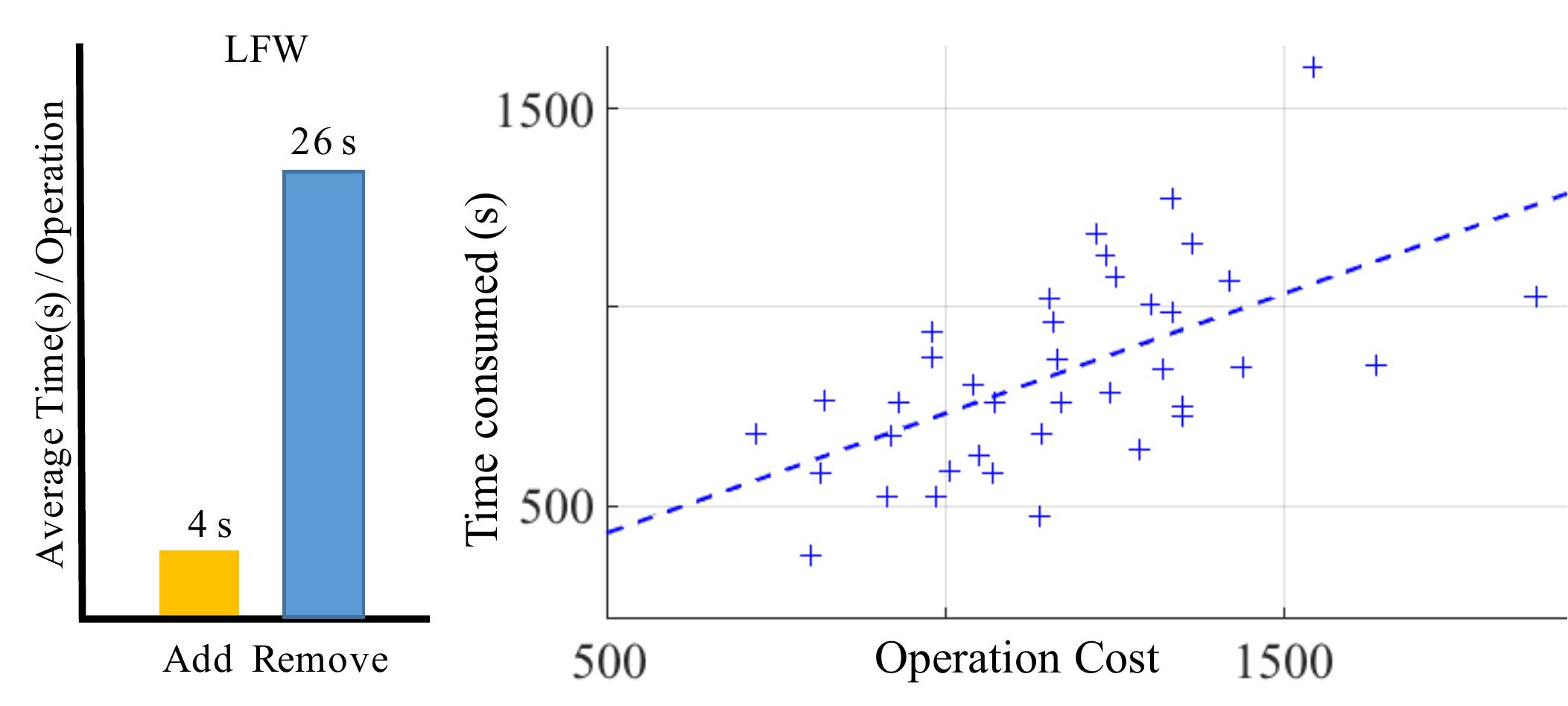}
\end{center}
\vskip -0.3cm
   \caption{A user study on operation cost.}
\label{fig:operationCost}
\end{figure}

The cost function $c(\cdot)$ can be obtained from a user study. In particular, we requested 30 volunteers and show them a number of randomly shuffled images as an album. Their task is to reorganize the photos into a desired groups' partition. We recorded the time needed for three types of operations: (1) adding a photo into a group, (2) removing a photo from a group, and (3) merging two groups. The key results are shown in Fig.~\ref{fig:operationCost}. It can be observed that the `removing' operation takes roughly 6$\times$ longer than the `adding' operation. The `merging' operation is almost similar to `adding'. Consequently, we set the cost for these three operations as 1, 6, 1, respectively. The validity is further confirmed by the plot in Fig.~\ref{fig:operationCost} that shows a high-correlation between the time consumed and the computed operation cost.

%\begin{figure}[t]
%    \centering
%    \begin{subfigure}[t]{0.7\linewidth}
%        \centering
%        \includegraphics[width=0.95\linewidth]{figure/operatorNumberTall.pdf}
%    \end{subfigure}%
%    ~ 
%    \begin{subfigure}[t]{0.3\linewidth}
%        \centering
%        \includegraphics[width=0.95\linewidth]{figure/AVGCost.pdf}
%    \end{subfigure}
%    \vskip -0.3cm
%    \caption{A user study on operation cost.}
%    \label{fig:operationCost}
%\end{figure}

%We refer the change of operator number between $K$ steps as our global reward $R_{\mathrm{OP}}$.

Given Eqn.~\eqref{eqn:operation_cost}, we define the long-term reward as:
\begin{equation}
\label{eqn:rewardLong}
R_\mathrm{long} = - \Delta Op^{(K)} = -(Op( h_{t-K} , g )-Op( h_t, g)),
\end{equation}
which encodes the operation cost changes in $K$ steps.

The key benefit brought by $R_\mathrm{long}$ is that it provides a long-term reward that guides an agent to thinking about the global picture of the grouping process. For any action that can hardly be decided (\eg, merging two noisy groups or merging a clean group with a noisy group), this term provides a strong evidence to the action-value function.  
%
%While IRL cannot guarantee a converge in an agent to exactly mimic an experts' behavior~\cite{abbeel2004apprenticeship}, the long-term reward, can help us to learn a better policy.

%------------------------------------------------------------------------------%
\subsection{Finding the Reward and Policy}
\label{subsec:learning}
%------------------------------------------------------------------------------%

%Let $T = \{t^{(i)}\}$ denotes a set of albums in a training set. The ground-truth partition for albums $t^{(i)}$ is given as $g^{(i)}$, and $\mathbf{X}^{(i)}$ is face representation. Our goal is learn a reward function and policy $\pi$ to grouping all these albums accurately.

As discussed in Sec.~\ref{sec:learning_face_grouping}, we assume the availability of a set training photo albums of which the ground-truth partition of the photos is known.
Let $\Omega = \{\omega^{(i)}\}$ denotes a set of albums in a training set. The ground-truth partition for albums $\omega^{(i)}$ is given as $g^{(i)}$, from which we can derive the ground-truth actions $\left\{a_\mathrm{GT}\right\}$ as an expert's behavior. Our goal is to find a reward function based on this behavior.
We perform the learning in two steps to ease the convergence of our method: (1) Firstly, we employ IRL~\cite{ng2000algorithms} to find the reward function with a myopic or short-sighted policy. (2) We then use the classic $\epsilon$-greedy algorithm~\cite{watkins1989learning} to find the optimal policy.

\begin{algorithm}[t]
\footnotesize{
\caption{Reward function learning via IRL.}
\label{alg:irl}
\SetAlgoLined
 \SetKwInOut{Input}{input}\SetKwInOut{Output}{output}
 \SetKwRepeat{Repeat}{repeat}{until}
 \Input{Training albums $\Omega = \{\omega^{(i)}\}$, ground-truth partition $\{g^{(i)}\}$}
 \Output{Binary classifier $(\mathbf{w},b)$ for $R_\mathrm{short}$ }
 Initialization $\mathbf{w} \leftarrow \mathbf{w}_0$, $b \leftarrow b_0$, $\mathcal{L} \leftarrow \emptyset$\;
 \Repeat{all albums are successfully partitioned}{
  \For{ $\omega^{(i)} \in \Omega$ }{
  	  $t = 0$\;
      Initialize partition $h_t$ with each photo as a single group\;
      Initialize history $H_t \leftarrow \emptyset$\;
      \Repeat{end of grouping}{
          $M$ recommends candidate groups $(C_j,C_k)$\;
          Compute action $a = \argmax_a R_\mathrm{short}(s,a)$\;
          Obtain ground-truth action $a_\mathrm{GT}$ based on $g^{(i)}$\;
          \If{$a \neq a_\mathrm{GT}$}{Add $(\phi(s), a_\mathrm{GT})$ into $\mathcal{L}$}
          \If{$a = a_\mathrm{GT}$}{$h_{t+1} \leftarrow \left\{ h_t \backslash \{C_j, C_k\} \right\} \cup \{ C_j \cup C_k \} $}
          Append $(C_j,C_k,a)$ into $H_{t+1}$\;
          $t=t+1$\;
      }
      Retrain $(\mathbf{w},b)$ on $\mathcal{L}$\;
  }
 }
 }
\end{algorithm}

\vspace{0.1cm}
\noindent
\textbf{Step 1}:
Algorithm~\ref{alg:irl} summarizes the first step. Specifically, we set $\gamma=0$ in Eqn.~\eqref{eqn:Qfunction} and $\beta=0$ in Eqn.~\eqref{eqn:reward}. This leads to a myopic policy $Q^\pi( s, a  | \pi ) $$= R_\mathrm{short}(s,a)$ that considers the current maximal short-term reward. This assumption greatly simplifies our optimization as $(\mathbf{w},b)$ of $R_{\mathrm{short}}$ (Eqn.~\eqref{eqn:rewardShort}) are the only parameters to be learned. We solve this using a binary RBF-kernel SVM with actions as the classes.
We start the learning process with an SVM of random weights and an empty training set $\mathcal{L}=$$\emptyset$. We execute the myopic policy repeatedly on albums. Once the agent chooses the wrong action w.r.t. the ground-truth, the representations of the involved groups and the associated ground-truth will be added to the SVM training set. Different albums constitute different games in which SVM will be continually optimized using the instances that it does not perform well. Note that the set $\mathcal{L}$ is accumulated, hence each time we use samples collected from over time for retraining $(\mathbf{w},b)$. The learning stops when all albums are correctly partitioned.

\vspace{0.1cm}
\noindent
\textbf{Step 2}:
Once the reward function is learned, finding the best policy $\pi$ becomes a classic RL problem. Here we apply the $\epsilon$-greedy algorithm~\cite{watkins1989learning}. $\epsilon$-greedy policy is a way of selecting random actions with uniform distribution from a set of available actions. Using this policy either we can select random action with $\epsilon$ probability and we can select an action with $1-\epsilon$ probability that gives maximum reward in a given state.
Specifically, we set $\gamma=0.9$ in Eqn.~\eqref{eqn:Qfunction} and $\beta=0.8$ in Eqn.~\eqref{eqn:reward}. We first approximate the action-value function $Q^\pi$ in Eqn.~\eqref{eqn:Qfunction} by a random forest regressor $Q(\phi(s),a)$~\cite{pyeatt2001decision}. The input to the regressor is $(\phi(s),a)$ and the output is the associated $Q^\pi$ value. The parameters of the regressor are initialized by $\phi(s)$, $a$, and $Q^\pi$ value, which are obtained in the first step (Algorithm~\ref{alg:irl}).
After the initialization, the agent selects and executes an action according to $Q$, \ie, $a = \argmax_a Q(\phi(s),a)$, but with a probability $\epsilon$ the agent will act randomly so as to discover a state that it has never visited before. %\footnote{The $\epsilon$ refers to the probability of random act in epsilon greedy policy}.
At the same time the parameters of $Q$ will be updated directly from the samples of experience drawn from the algorithm's past games. At the end of learning, the value of $\epsilon$ is decayed to 0, and $Q$ is used as our action-value function for policy $\pi$.
%\cite{hitesh2010fuzzy}
%We refer readers to~\cite{sutton1988learning} for a systematic introduction to Temporal difference (TD-$\lambda$) algorithm.

%\subsection{Noise Face and Post-processing}
%
%One big challenge is that the profiles and false detected false faces are often near to each other at feature space. Here we make two modification on our framework to further attack noisy samples. 
%
%\begin{enumerate}
%  \item In training phase, if an agent tend to merge 2 noise face, or noise face with unnamed group (group that no ones face dominate more than half faces ). We didn't count it as an incorrect action.
%  \item After clustering, we compute quality attribute on all faces use the quality classifier we trained (in feature representation section). If one cluster has more than 60\% low quality faces, we removed it. This post-processing was done for every experiments we tested, we also report the result without post-processing in supplementary material. 
%\end{enumerate} 

%%%%%%%%%%%%%%%%%%%%%%%%%%%%%%%%%%%%%%%%%%%%%%%%%%%%%%%%%%%%%%%%%%%%%%%%%%%%%%%%%
\section{Experiments}
\label{sec:experiments}
%%%%%%%%%%%%%%%%%%%%%%%%%%%%%%%%%%%%%%%%%%%%%%%%%%%%%%%%%%%%%%%%%%%%%%%%%%%%%%%%%

\noindent
\textbf{Training Data:}
Our algorithm needs to learn a grouping policy from a training set. The learned policy can be applied to other datasets for face grouping. Here we employ $2,000$ albums simulated from MS-Celeb-1M~\cite{guo2016msceleb} of 80k identities as our training source.
We will release the training data. 

\begin{figure}[t]
\begin{center}
   \includegraphics[width=0.98\linewidth]{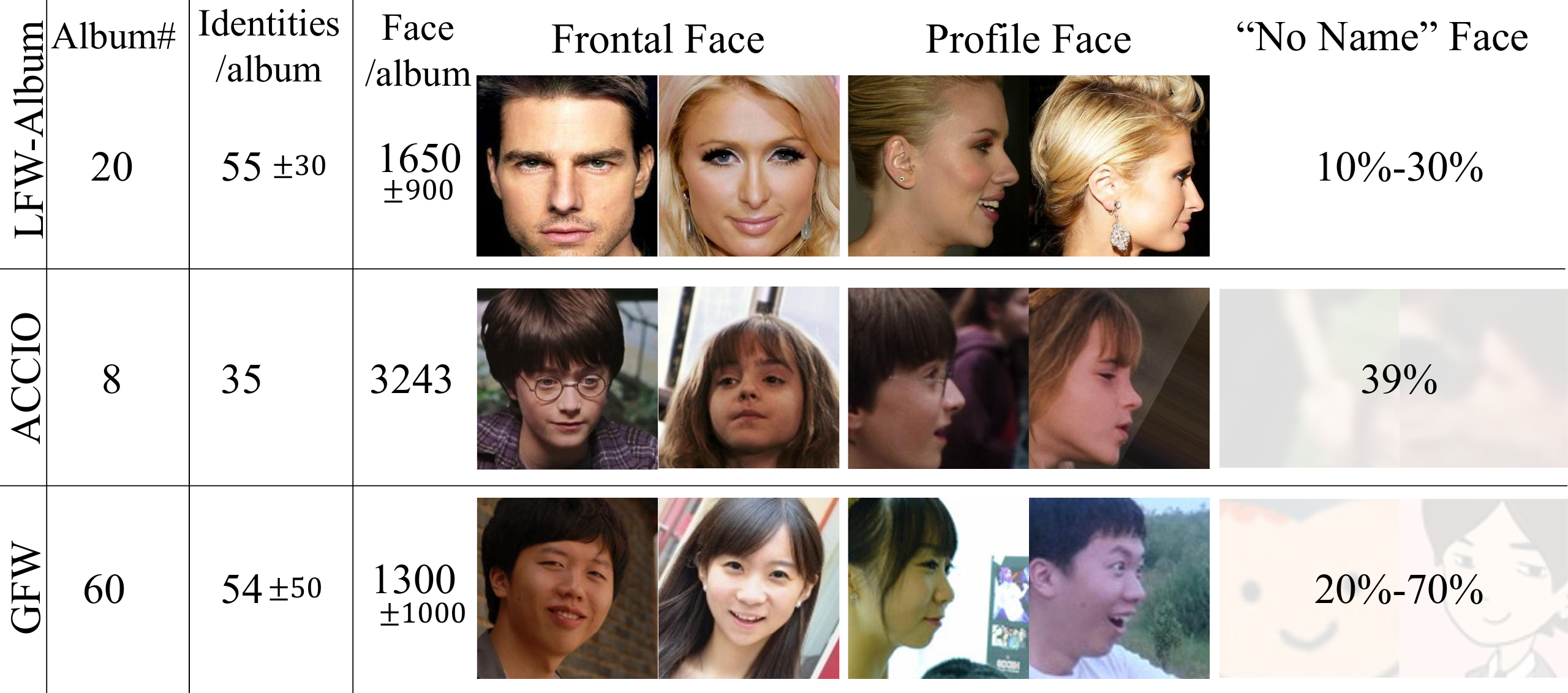}
\end{center}
\vskip -0.5cm
   \caption{Overview of test datasets.}
\label{fig:datasetOverview}
\end{figure}

\noindent
\textbf{Test Data:} To show the generalizability of the learned policy, we evaluate the proposed approach on three datasets of different scenarios exclusive from the training source. Example images are provided in Fig.~\ref{fig:datasetOverview}.

\noindent
\textit{1) LFW-Album}: We construct a challenging simulated albums from LFW~\cite{huang2007labeled}, MS-Celeb-1M~\cite{guo2016msceleb}, and PFW~\cite{Sengupta2016frontal}, with a good mix of frontal, profile, and non-interested faces. We prepare 20 albums and with exclusive identities. Note that the MS-Celeb-1M samples used here are exclusive from the training data. 

\noindent
\textit{2) ACCIO Dataset}: This dataset~\cite{esam2015accio} is commonly used in the studies of video face clustering. It contains face tracklets extracted from series of Harry Potter movie. Following~\cite{zhang2016joint}, we conduct experiments on the first instalment of the series, which contains 3243 tracklets from 36 known identities. For a fair comparison, we do not consider uninterested faces in this dataset following~\cite{zhang2016joint}. We discard the temporal information and used only the frames in our experiments.

\noindent
\textit{3) Grouping Face in the Wild (GFW)}: To better evaluate our algorithm for real-world application, we collect 60 real users' albums with permission from a Chinese social network portal. The size of an album varies from 120 to 3600 faces, with a maximum number of identities of 321. In total, the dataset contains 84,200 images with 78,000 faces of 3,132 different identities. All faces are automatically detected using Faster-RCNN~\cite{ren2015faster}. False detections are observed. We annotate all detections with identity/noise labels.
The images are unconstrained, taken in various indoor/outdoor scenes. Faces are naturally distributed with different poses with spontaneous expression. In addition, faces can be severely occluded, blurred with motion, and differently illuminated under different scenes. We will release the data and annotations. To our knowledge, this is the largest real-world face clustering dataset.

Given the limited space, we exclude results on traditional grouping datasets like Yale-B~\cite{georghiades2001few, lee2005acquiring}, MSRA-A~\cite{zhu2011rank}, MSRA-B~\cite{zhu2011rank} and Easyalbum~\cite{cui2007easyalbum}. Yale-B were captured in controlled condition with very few profile faces and noises. The number of albums is limited in the other three datasets. %Nonetheless, we still conducted experiments on \textbf{xx} and \textbf{xx}. We refer readers to our supplementary material for the results.

\begin{table*}[t]

\footnotesize

\begin{center}

\caption{Face grouping results on LFW-Album, ACCIO-1, and GFW.}

%\vskip -0.25cm

\begin{tabular}{l|ccc|c||ccc|c||ccc|c}

Dataset & \multicolumn{4}{|c||}{LFW-Album} & \multicolumn{4}{|c||}{ACCIO-1} & \multicolumn{4}{|c}{GFW} \\ \hline

Metric & P(\%) & R(\%) & $F_1$(\%) & $\overline{Op}$  & P(\%) & R(\%) & $F_1$(\%) & $\overline{Op}$ & P(\%) & R(\%) & $F_1$(\%) & $\overline{Op}$ \\

\hline \hline

K-means & 73.6& 86.6& 79.3& 1.12  & 72.2& 34.4& 46.6& 0.65  & 66.6& 35.7& 41.1& 1.47  \\

   GDL~\cite{zhang2012graph}   & 66.5& \textbf{92.2}& 76.4& 1.21  & 18.1& 91.1& 30.2& 3.51  & 67.4& 59.4& 55.9& 1.30  \\

   HC    & 74.2& 80.8& 76.6& 0.35  & 17.1& \textbf{91.9}& 28.9& 3.28  & 77.5& 22.3& 15.0& 0.81  \\

   AP~\cite{brendan2007clustering}    & 76.7& 71.1& 73.7& 1.07  & 82.2& 9.6& 17.1& 0.59  & 69.7& 25.3& 32.7& 0.86  \\

   Deep Adaptation~\cite{zhang2016joint} & - & - & - & - & 71.1& 35.2& 47.1& - & - & - & - & - \\ \hline \hline

   IL-Kmeans & 76.7& 87.8& 81.6& 0.95  & 82.8& 34.1& 48.3& 0.54  & 53.4& 43.6& 43.3& 1.17  \\ 

   IL-GDL & 79.9& 90.1& 84.5& 0.54  & 88.6& 46.3& 60.8& 0.78  & 78.4& \textbf{76.2}& \textbf{74.5} & 0.68  \\

   %IL-HC & \textbf{96.7}& 85.2& \textbf{90.6}& \textbf{0.14}  & \textbf{90.8}& 78.6& \textbf{84.3}& \textbf{0.52}  & \textbf{85.0}& 74.3& \textbf{76.0}& \textbf{0.39}  \\

   IL-HC & \textbf{97.8} & 85.3& \textbf{91.1}& \textbf{0.14}  &  \textbf{90.8}& 78.6& \textbf{84.3}& \textbf{0.52} & \textbf{96.6} & 53.7& 67.3& \textbf{0.17}  \\ \hline \hline

 %  Recomm.-Independent & 96.5 & 52.4 & 66.5 & 0.17  & \textbf{92.2} & 89.0 & \textbf{90.6} & \textbf{0.21}  & 90.7 & 75.9 & \textbf{82.7} & 0.55 \\

   SVM $+$ Deep Features & 82.7 & 87.4 & 85.0 & 0.45& 89.0 & 61.3 & 72.6 & 0.74 & 84.3 & 46.4 & 56.3 & 0.33 \\
   Siamese Network $+$ Deep Features & 87.1 & 87.6 & 87.3 & 0.44 & 59.7 & 88.1 & 71.2 & 0.79 & 49.9 & 92.3 & 62.8 & 0.33 \\\hline

\end{tabular}

\label{tab:mainExperiment}

\end{center}

\vskip -0.6cm

\end{table*}

\noindent
\textbf{Implementation Details:}
All face representation are extracted from Inception-v3 model~\cite{szegedy2015rethinking} fine-tuned with MS-Celeb-1M~\cite{guo2016msceleb}. 
We suggest some parameter settings as follows. We set $\beta=0.8$ in Eqn.~\eqref{eqn:reward} to balance the scales of short- and long-term rewards. We fixed the number of faces $\eta=5$ to form the similarity and quality features (Sec.~\ref{subsec:short_term}). The five shortest distances is a good trade-off between performance and feature complexity. If a group has fewer than five faces (to the extreme only one face exists), we pad the distance vector with the farthest distance.

\noindent
\textbf{Evaluation Metrics:}
We employ multiple metrics to evaluate the face grouping performance, including the B-cubed precision, recall, and $F_1$ score suggested by ~\cite{zhang2013unified} and~\cite{zhang2016joint}. Specifically, B-cubed recall measures the average fraction of face pairs belonging to the ground truth identity assigned to the same cluster. And B-cubed precision is the fraction of face pairs assigned to a cluster with matching identity labels. The $F_1$ score measures the harmonic means of these two metrics.
We also use \textsl{operation cost} introduced in Sec.~\ref{subsec:long_term}. To facilitate comparisons across datasets of different sizes, we compute the operation cost normalized by the number of photos as our metric, \ie, $\overline{Op} = Op/ N$. We believe that this metric is more important than the others since it directly reflects how much effort per image a user needs to spend to organize a photo album.

%------------------------------------------------------------------------------%
\subsection{Comparison with Unsupervised Methods} 
\label{subsec:comparison_exp}
%------------------------------------------------------------------------------%

We compare our method with classic and popular clustering approaches: 1) K-means, 2) Graph Degree Linkage (GDL)~\cite{zhang2012graph}, 3) Hierarchical Clustering (HC), and 4) Affinity Propagation (AP)~\cite{brendan2007clustering}. Note that we also compare with~\cite{zhang2016joint}. Since the code is not publicly available, we only compare with its reported precision, recall, and $F_1$ scores on the ACCIO-1 dataset. 
Note that these baselines use the same features as our approach, as discussed in Sec.~\ref{subsec:short_term}.
To verify if the proposed imitation learning (IL) framework helps existing clustering methods, we adapt K-means, GDL and HC into IL-K-means\footnote{For IL-K-means algorithm, the action space $A$ is no longer binary due to the nature of K-means. Here we adapt the framework to have an action space of $K+1$, for determining the merging of a sample into one of the $K$ clusters. And we replace the SVM with a RankSVM~\cite{Joachims2002optimizing} to compute the rewards for each cluster.}, IL-GDL and IL-HC to equip them with the sequential decision capability. This is achieved by using the respective algorithm as the recommender (see Sec.~\ref{sec:overview}).

Table~\ref{tab:mainExperiment} summarizes the results on three datasets. We observed that: (1) imitation learning consistently improves the different clustering baselines. For instance, on LFW-Album, the $F_1$ score and $\overline{Op}$ of HC improves from 76.6\% and 0.35 to 91.1\% to 0.14. Notably, IL-HC outperforms other variants based on the proposed IL, although our framework is not specifically developed to work only with hierarchical clustering.
(2) The operation cost is lower with a high-precision algorithm. This result matches with our user study since a user is good at adding similar photos into a group but poor at removing noisy faces that can be hard to distinguish.
% Later we will show that precision and recall of an algorithm can be adjusted by manipulating the cost distribution.

\begin{figure}[t]
\begin{center}
\includegraphics[width=1\linewidth]{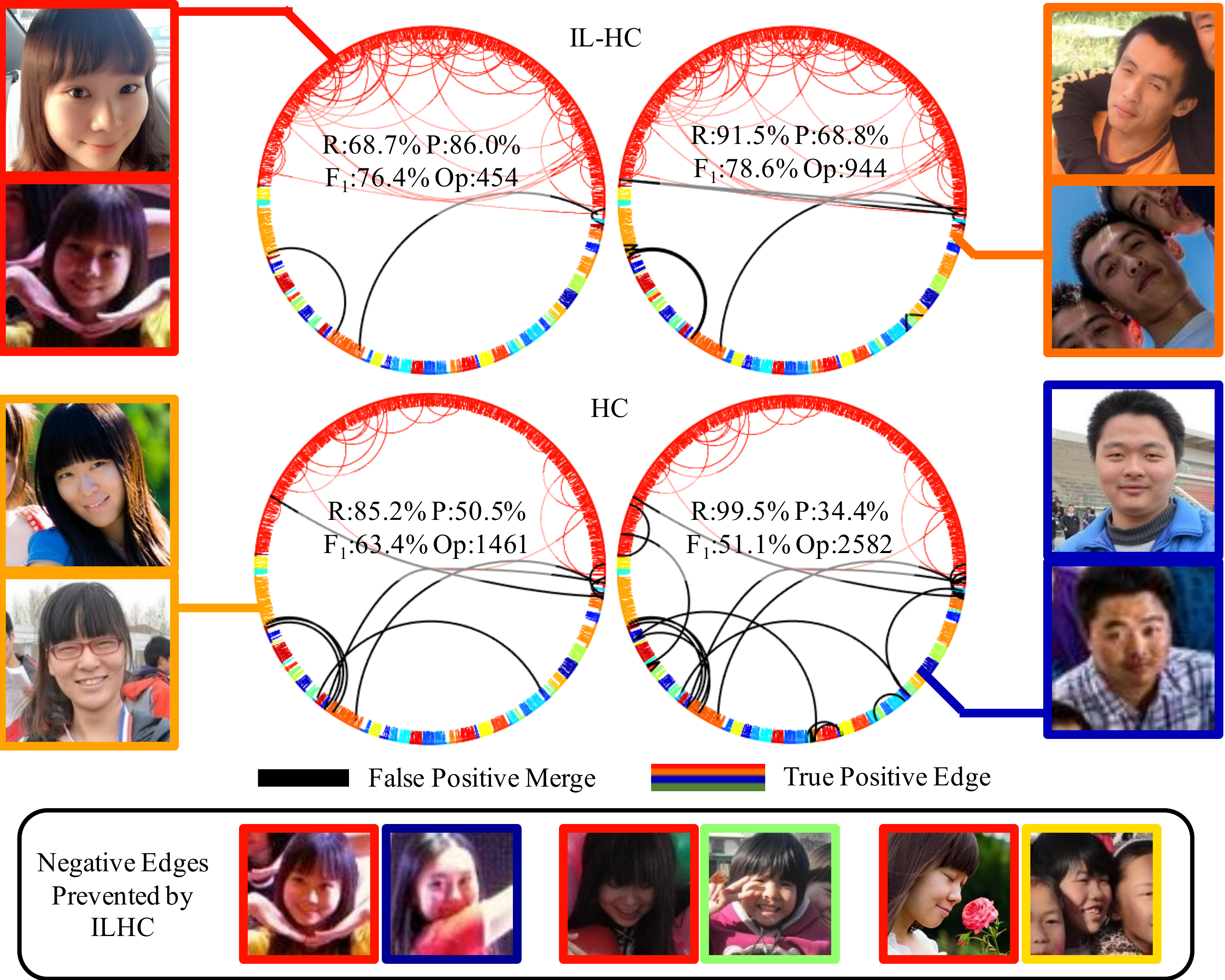}
\caption{\textbf{Visualization of grouping result} (best viewed in color). The figure shows a typical album of GFW. Each node in the circle indicates a face, and each color denotes a person of specific identity. The result of hierarchical clustering (HC) contains more negative edges (black arcs), representing wrongly merged groups. Imitation learning based HC (IL-HC) prevents negative edges as produced by HC.}
\label{fig:example_grouping}
\end{center}
\vskip -0.4cm
\end{figure}

We compare grouping results of IL-HC and HC qualitatively in Fig.~\ref{fig:example_grouping}. IL-HC yields more coherent face groupings with exceptional robustness to outliers.

%------------------------------------------------------------------------------%
\subsection{Comparison with Supervised Methods}
\label{subsec:comparison_supervised}
%------------------------------------------------------------------------------%

%Our framework can be regarded as a supervised approach since it learns from experts' demonstrations. 
We compare our framework with two supervised baselines, namely a SVM classifier and a three-layer Siamese network. The three layers of the Siamese network have 256, 64, 64 hidden neurons, respectively. A contrastive loss is used for training. To train the baselines, each time we sample two subsets of identities from MS-Celeb-1M as the training data. SVM and the Siamese Network are used to predict if two groups should be merged or not.
Features are extracted following the method presented in Sec.~\ref{subsec:short_term}. These supervised baselines are thus strong since their input features are identical to those we use in our IL framework. The features include face similarity vector that is derived from Inception-v3 face recognition model fine-tuned with MS-Celeb-1M dataset. The deep representation achieves 99.27\% on LFW, which is better than~\cite{sun2014deep} and on-par with~\cite{wen2016discriminative}.  
The results of the baseline are presented in Table~\ref{tab:mainExperiment}. It is observed that the IL-based approach outperforms the supervised baselines by a considerable margin.

%------------------------------------------------------------------------------%
\subsection{Ablation Study}
\label{subsec:ablation}
%------------------------------------------------------------------------------%

\begin{figure}[t]
\begin{center}
   \includegraphics[width=1.0\linewidth]{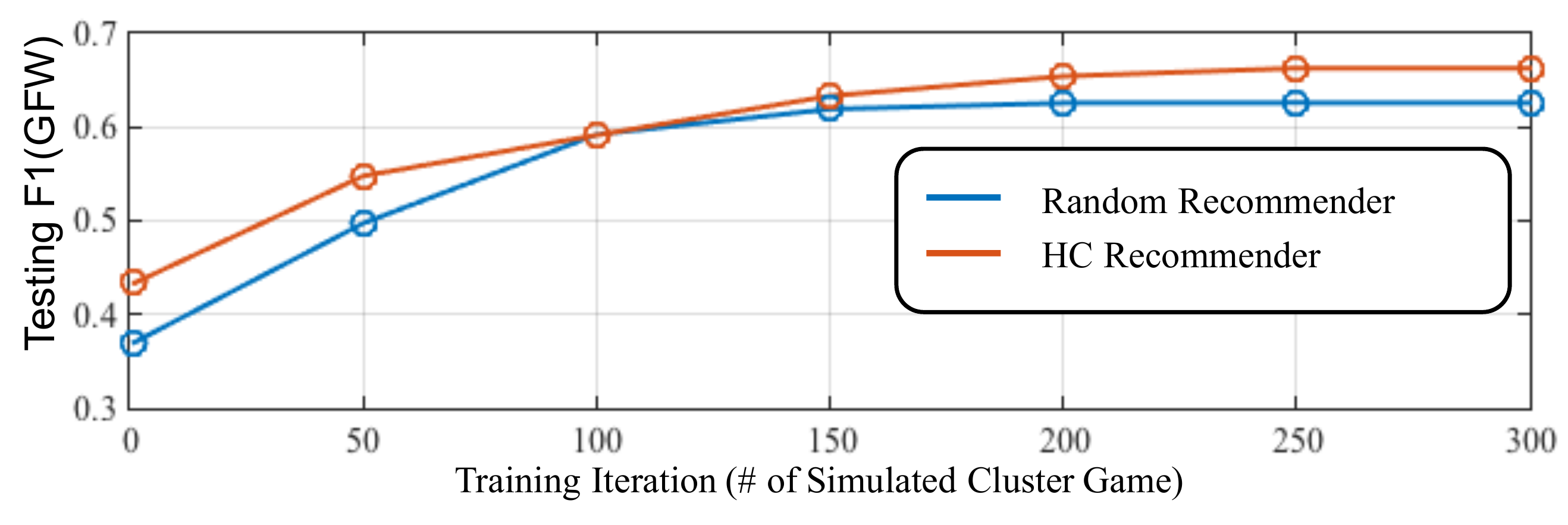}
\end{center}
\vskip -0.50cm
   \caption{The $F_1$ score on using different recommenders along with different training iterations. The red curve is obtained by using Hierarchical Clustering (HC) as the recommender, while the blue curve is obtained by using the random recommender.}
\label{fig:randomRecommender}
\vskip -0.2cm
\end{figure}

\noindent
\textbf{Further Analysis on Recommender}:
In Sec.~\ref{subsec:comparison_exp}, we tested three different recommenders based on different clustering methods, namely K-means, GDL, and HC.
In this experiment, we further analyze the use of a random recommender that randomly chooses a pair to recommend. 
%
%Figure~\ref{fig:randomRecommender}(a) shows its iteration-error curves. In comparison to the recommender based on hierarchical clustering, which always recommends the nearest groups, the random recommender exhibits a slower convergence and poorer results.  
% 
%As observed in Figure~\ref{fig:randomRecommender}(b), it is worth pointing out that our method with the random recommender still achieves a $F_1$ score of 61.9\% on GFW, which outperforms the unsupervised baseline, which only achieves 15\%. The results suggest the robustness of our method against different types of recommenders.
%
Figure~\ref{fig:randomRecommender} shows the $F_1$ score comparisons between a Hierarchical Clustering (HC) recommender and a random recommender. In comparison to the recommender based on HC, which always recommends the nearest groups, the random recommender exhibits a slower convergence and poorer results. It is worth pointing out that the random recommender still achieves a $F_1$ score of 61.9\% on GFW, which outperforms the unsupervised baseline, which only achieves 15\%. The results suggest the usefulness of deploying a recommender.

We also evaluate an extreme approach that does not employ a recommender but selects a group pair to merge based on the values produced by the learned action-value function. Specifically, in each step, we compute the $Q(\phi(s),a)$ exhaustively for all possible pairs of group, and select the pair with the highest value to merge. This approach achieves $82.7\%$ $F_1$ on GFW. It is not surprising that the result is better than our IL-HC as this approach performs exhaustive search for pairs. This method has a runtime complexity of $\BigO(N^3)$, much higher than the IL-HC. The results suggest the effectiveness of the clustering-based recommender in our framework.

\noindent
\textbf{Discard the Face Quality Feature}: 
If we remove the face quality feature from the feature vector $\phi(s)$, the $F_1$ score achieved by IL-HC of LFW-Album, ACCIO-1, and GFW will drop from 91.1\%, 84.3\%, and 67.3\%, to 89.5\%, 65.0\%, and 48.4\%, respectively.
The results suggest that the importance of quality measure depends on the dataset. Face quality feature is essential on the GFW dataset but less so on others, since GFW consists more poor-quality images.

\noindent
\textbf{Reward Function Settings}:
We evaluate the effect of two reward terms in the reward function defined in Eqn.~\eqref{eqn:reward}.

\noindent
1) $R_{\mathrm{short}}$ \& $R_{long}$: The full reward setting with $\beta \neq 0$.

\noindent
2) w/o $R_\mathrm{long}$: Without the long-term reward based on operation cost, \ie, $\beta = 0$.

\noindent
3) w/o $R_\mathrm{short}$: In this setting, we discarded $R_{\mathrm{short}}$ learned by IRL, and redefined it to take a na\"{i}ve $\pm 1$ loss, \ie, $R_{\mathrm{short}} = \mathbbm{1}(a = a_\mathrm{GT} )$, where $\mathbbm{1}(\cdot)$ is an indicator function that outputs 1 if the condition is true, and -1 if it is false. %Specifically, $R_{\mathrm{short}}=1$ if an action $a$ is identical to the ground-truth $a_\mathrm{GT}$, and $-1$ otherwise. The reward thus turns into a classic reward function. We set $\beta \neq 0$.

\begin{figure}[t]
\begin{center}
   \includegraphics[width=0.95\linewidth]{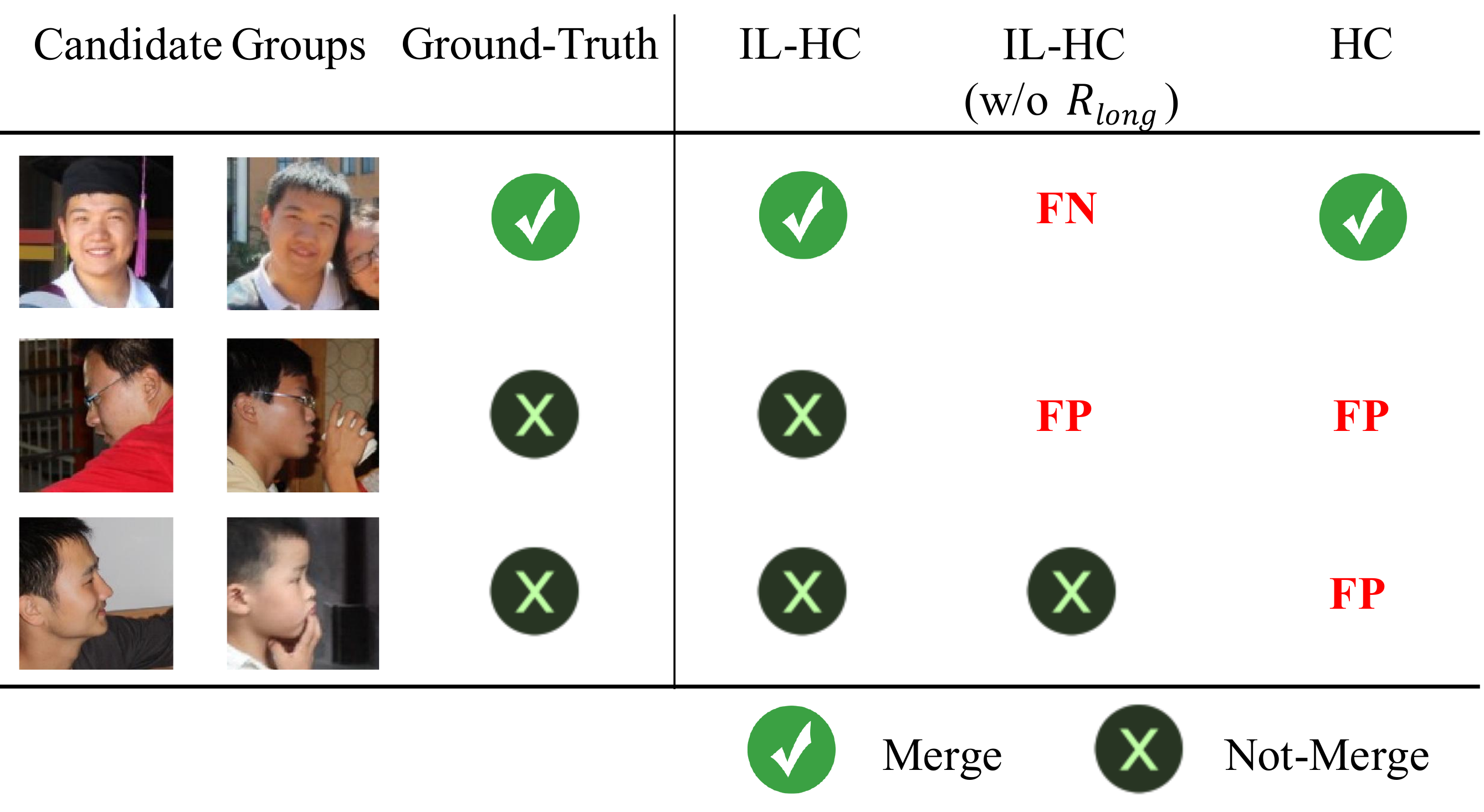}
\end{center}
\vskip -0.4cm
   \caption{Example of merging decisions made by different algorithms. Each image represents a group they belong to. It is observed that IL-HC w/o $R_{long}$ tends to produce false negative (FN) and false positive (FP) mistakes in comparison to IL-HC with full reward.}
\label{fig:qualitative}
\end{figure}

\begin{figure}[t]
\begin{center}
   \includegraphics[width=0.95\linewidth]{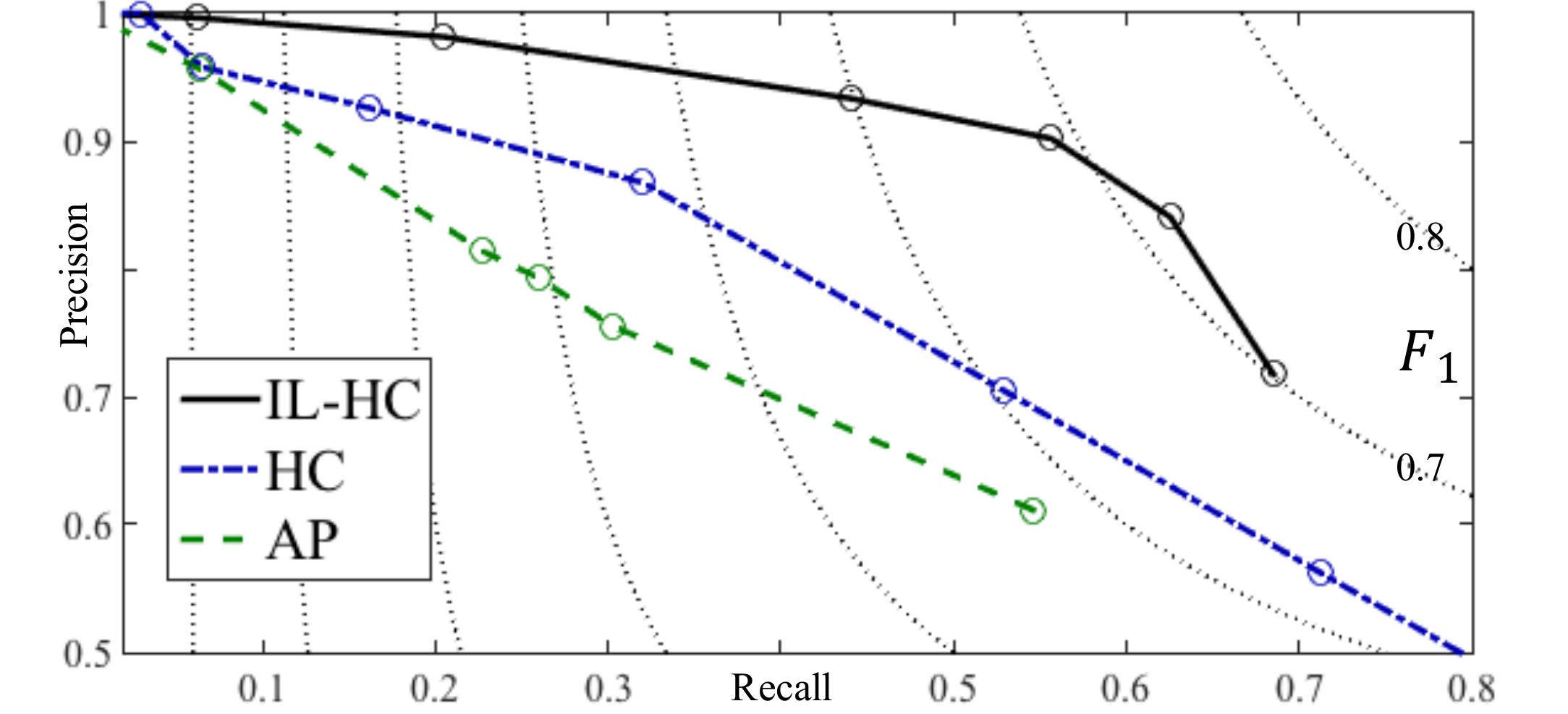}
\end{center}
\vskip -0.4cm
   \caption{B-cubed precision and recall curve on GFW dataset with adjustment to the cost distribution of `add, remove, merge' operations in $R_\mathrm{long}$.}
\label{fig:PRcurve}
\end{figure}

The results reported in Table~\ref{tab:valueExperiment} shows that both short- and long-term rewards are indispensable to achieve good results. 
Comparing the baselines ``w/o $R_\mathrm{short}$'' against the full reward, we observed that IL learned a more powerful short-term reward function than the na\"{i}ve $\pm 1$ loss.
Comparing the baselines ``w/o $R_\mathrm{long}$'' against the full reward, albeit removing $R_\mathrm{long}$ only reduces the $F_1$ score slightly, the number of false positive and false negative merges actually increase for noisy and hard cases.
Figure~\ref{fig:qualitative} shows some representative groups that were mistakenly handled by IL-HC w/o $R_\mathrm{long}$.
It is worth pointing out that by adjusting the cost distributions of $R_\mathrm{long}$, \eg, changing the cost of `add, remove, merge' from (1,6,1) to (1,1,1), one could alter the algorithm's bias on precision and recall to suit for different application scenarios. A chart of B-cubed PR-curves is depicted in Fig.~\ref{fig:PRcurve} to show the influence of cost distribution. Hierarchical clustering with imitation learning (IL-HC) outperforms the baselines HC and AP no matter which settings we use. We recommend a high precision setting in order to achieve a low normalized operation cost $\overline{Op}$, as suggested by experiments in Sec.~\ref{subsec:comparison_exp}.

\begin{table}[t]
\footnotesize
\begin{center}
\caption{Different settings of reward function. We use IL-HC in this experiment.}
%\vskip -0.2cm
\begin{tabular}{l|cc|cc}
  Dataset & \multicolumn{2}{|c|}{LFW-Album} & \multicolumn{2}{|c}{GFW} \\ \hline
  Metric &  $F_1$(\%) & $\overline{Op}$  & $F_1$(\%) & $\overline{Op}$ \\
  \hline \hline
  $R_{\mathrm{short}}$ \& $R_\mathrm{long}$ & 91.1 & 0.14 & 67.3 & 0.17  \\ 
  w/o $R_{long}$  & 90.7 & 0.14 & 62.6 & 0.17  \\
  w/o $R_\mathrm{short}$ & 73.0 & 0.54 & 17.1 & 0.65  \\ \hline
\end{tabular}
\label{tab:valueExperiment}
\end{center}
\end{table}

\section{Conclusion}
%%%%%%%%%%%%%%%%%%%%%%%%%%%%%%%%%%%%%%%%%%%%%%%%%%%%%%%%%%%%%%%%%%%%%%%%%%%%%%%%%

We have proposed a novel face grouping framework that makes sequential merging decision based on short- and long-term rewards. With inverse reinforcement learning, we learn powerful reward function to cope with real-world grouping tasks with unconstrained face poses, illumination, occlusion, and abundant of uninteresting faces and false detections. We have demonstrated that the framework benefits many existing agglomerative-based clustering algorithms.

%\cavan{ need to include references pointed out by reviewers.}  

{\small
\bibliographystyle{ieee}
\bibliography{short,egbib}
}

\end{document}